\documentclass[cameraready]{Interspeech}

\title{LLMs and the ZPD}

\author[affiliation={1}]{Peter}{Wallis}

\address{
    $^1$ Centre for Policy Modelling, UK
}

\email{pwallis@acm.org}


\usepackage{comment}
\usepackage{changepage}
\usepackage{soul}
\newenvironment{quote2c}{\begin{adjustwidth}{5mm}{}}{\end{adjustwidth}}

\begin{document}

\maketitle

\begin{abstract}
One hundred years ago Vygotsky and his circle were exploring the nature of consciousness and defining what would become psychology in the Soviet Union.  They concluded that children develop ``scientific thinking'' through interacting with enculturated adults in Zones of Proximal Development or ZPDs. The proposal is that, contrary to the claims of some, the LLM mechanism is not doing thinking with ``distributed representations,'' but rather the completion model is doing ``primitive thinking'' in terms of \textit{practices}.  Viewed from this perspective, it would seem our large language models don't hallucinate, but rather dream, and that what is needed is not ``guard rails'' but an investigation of the set of cognitive tools that enable us to do things that look like common-sense.  The proposal here is that \textit{interaction} is core to human communication rather than just an add-on to ``real'' understanding.
\end{abstract}

\section{Introduction}

In an article in The New Yorker~\cite{GAC} Geoffrey Hinton is worried that Artificial Neural Networks, with the benefit of massive training data, have unleashed Artificial General Intelligence (AGI) on the world.  His counter to the doubters is quoted at length as follows:
\begin{quote2c}
  People say, ``It's just glorified autocomplete,''... Now, let's analyze that.  Suppose you want to be really good at predicting the next word.  If you want to be \textit{really} good, you have to understand what's being said.  That's the only way.  So by training something to be really good at predicting the next word, you're actually forcing it to understand.  Yes, it's 'autocomplete' -- but you didn't think through what it means to have really good autocomplete.  [The New Yorker, November 13, 2023]
\end{quote2c}
Hinton believes that large language models can comprehend the meaning of words and have ideas, but others obviously think the mechanism is more limited.
Thinking about thinking is not new and one hundred years ago Vygotsky and his co-workers were tasked with educating the masses. In the process they shaped Soviet psychology and pointed out that: 
\begin{quote2c}
  In ontogenesis one can discern a preintellectual stage in the development of speech, and a prelinguistic stage in the development of thought.
\end{quote2c}
Vygotsky goes on:
\begin{quote2c}
  At a certain moment these two developmental lines become intertwined, whereupon thought becomes verbal, and speech intellectual. This moment signifies a switch from a natural track of development to a cultural one. [Kozulin~\cite{kozulin90},p153]
\end{quote2c}
This switch in a child's thinking happens over time, only once the child has the ``cognitive tools'' to progress, and only in interaction with adults.  The child learns to think as an adult in Zones of Proximal Development or ZPDs.

\begin{figure}[t]
  \centering
  \includegraphics[width=\linewidth]{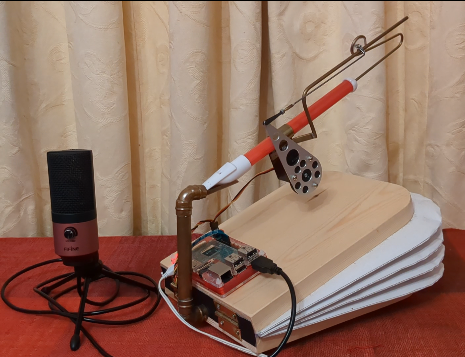}
  \caption{A swanee whistle, bellows, and a microphone for an ethnographic study of Levinson's 'interaction engine' and pre-intellectual speech.}
  \label{whistle}
\end{figure}

The proposal is that the completion model of Large Language Models is actually a good model of ``primitive thinking'' and that thinking with symbols and representation is something LLMs need to be taught.  This may seem a strange claim given language is our archetype symbol system and ``language'' is in the name, but I will argue they are using language \textit{performatively} rather than exploiting the ability of natural language to represent things.
Using speech performatively -- using vocalisations to cause others to do things -- is a study of people.  It is a study of the community of practice of the speech users.  Figure~\ref{whistle} is of a device that interacts with people, but without words. It cannot use speech "scientifically," and Word Error Rate is simply not applicable.  So how should we describe the observations we make?

The next section looks at prelinguistic thought, and Section~\ref{s3} at pre-intellectual speech. Section~~\ref{s4} combines the two and Section~\ref{s5} outlines two approaches to studying vocal communication without words.

\section{Natural Intelligence} \label{s2}

Hinton's position is a common one based on a long history of thinking about thinking with symbols and representation. In classic AI this was explicit with symbols such as BRICK\_3 representing an actual brick in the world, and when the bricks were on a table top, this type of reasoning worked even on the computers available in the 1970s.  A symbol such as ON might represent a relationship between brick 3 and brick 2 and so the statement ON(BRICK\_2,BRICK\_3) could be tested for truth, and planning could be done in terms of axioms and rules.  This model of thinking turns out to be fragile, and it seemed we'd need an awfully large amount of knowledge about the world in order to make machines with even an approximation of common-sense.

By 1990 there was generally a consensus that something was wrong with the approach and Harnad wrote his famous paper ``The Symbol Grounding Problem''~\cite{har90} which concluded that (artificial) neural nets were the way forward because they use ``distributed representations''.  Others, famously Brooks in his paper ``Intelligence without Representation''~\cite{brooks91}, were taking a different approach and exploring ``embodied intelligence''.  Hinton, surfing in on a wave of interest in neural nets, is taking Harnad's line and saying real thinking is done with symbols, but we just can't point to where the symbols are. The argument here is in favour of the embodiment approach.  Embodiment is often seen as involving a body (obviously) but embodiment is perhaps better seen as an issue of \textit{situated action}.  Meaning is grounded, not so much in the body~\cite{ruthrof97}, but in the relationship between the agent and its operating environment~\cite{Agre88,Mill89,HeJa96}. In Section~\ref{s3} we consider ChatGPT as an agent situated in a social world but first, what can be done with a glorified autocomplete mechanism?

\subsection{Glorified Autocomplete -- more useful than it looks}

Artificial Neural Networks are notoriously opaque and the relevant research community seem to have evolved their own way of talking about progress.  There are plenty of ``explainer'' YouTube videos, and attention (a term with strong affiliation with consciousness) seems to have morphed into meaning bag-of-words on a span of input (context). The language is about the mechanism, not about what it does, although the allusions come thick and fast.  There is a long history of having statistical models of a training set from a population of samples. The resultant model can then be used to produce original samples that have a fit with the population. Markov chains are a classic, and I believe there is something known as multi dimensional kernels that provide a suitably theoretical view of the autocomplete problem.

For the rest of us, we tend to think LLMs are thinking much as we do in terms of things and their relationships.  However consider an LLM playing chess~\cite{chess,more-chess}. Being good at chess is classic indicator of intelligence and was used in AI research from the start.  As described in the Dynomight blog, LLMs are not consistently good at chess, nor indeed consistently not completely rubbish. When an LLM is successfully prompted to play chess (and the tokenisation is fine tuned) the LLM can be quite good.  But is it building a cognitive model of the game?  An alternate interpretation of the behaviour is that the machine is running (glorified) autocomplete over the beginning of a chess game and, with a bit of framing to take turns and stick to the rules, it can \textit{complete} a good game of chess.  Following Hinton and Harnad, the machine might be dealing with ``distributed representations'' of boards and pieces and reasoning ``scientifically'', but looking at the tokens used in the Dynomight analysis, they are things like:  1. e4 e5 2. Nf3 Nc6 3. ... and although ``Nf3'' might be seen as a reference to a chess piece, is it the same piece as ``Nc6''?  Well it is the same physical object at a different time and in a different place. Recognising this cannot be done by looking at the tokens alone.  The claim here is that the token sequence the system is completing is representations of \textit{moves}.  The thing being completed is the \textit{performance} of a (good) chess game.  It might be the case that the system could play a better game of chess if it was reasoning about chess pieces and had a theory of mind of its opponent, but being able to reproduce all the chess games on the internet is, it seems, also a way to play good chess.

\begin{figure}[h]
  \centering
  \includegraphics[width=\linewidth]{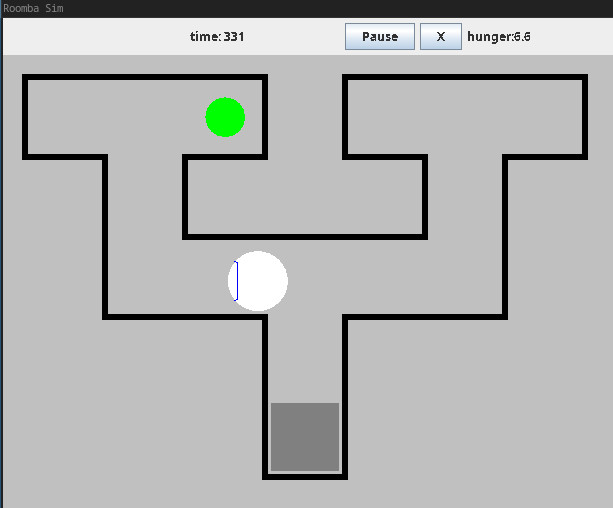}
  \caption{``just glorified autocomplete'' as a model of Vygotsky's ``natural'' intelligence.}
  \label{roombasim}
\end{figure}

So gpt-3.5-turbo-instruct does play chess, and we have an explanation in terms of autocomplete.  The interest here however is to convince the reader that glorified auto complete is able to do what the received wisdom tells us requires reasoning. In psychology mazes are used to investigate the inner workings of rats' minds.   In an unpublished paper I describe a simulation of a Roomba running a maze (see Figure~\ref{roombasim}).
Whereas psychologists hypothesize about what is happening in a rat brain based on its behaviour, we can actually look inside the robot and see how the behaviour is generated.
My (simulated) robot has an executive function that reasons over \textit{behaviours} rather than actions.
A behaviour is a reflexive unit that, from a teleological perspective, fulfills a need.
As a situated agent, there is a relationship between this agent and the environment in which it successfully operates.  And rather than carving the world into things, this relationship is carved into a set of discrete behaviours.
The executive function - the bit which does thinking - does not think about how each behaviour works, just what it can do.  The terminology is from Arkin~\cite{Ark98} but the argument is from the enactivists~\cite{enactivism17}. The executive function, rather than reasoning about goals and beliefs and formulating intentions~\cite{bip}, uses case-based reasoning~\cite{cbr} to choose what to do next based on what happened last time.  Given a sequence of tokens representing the enablement and activation of different behaviours, the system predicts what will happen next by ``autocompleting'' recent events.

Consider two views of the real world practice of going to the supermarket.  As an encultured Western adult I would describe my route as turning left at the phone repair shop, and then finding the Aldi on the left about a kilometer up.  There are \textit{things} to see and I place them and myself on a mental representation of a map.  The description for my teenager would need to be different.  I might say go to the place where we had your phone repaired, then go up the street next to that and follow the road. You will come across a sign that indicates you can do Aldi shopping here. Like GPT-3.5 turbo playing chess, the cases are in terms of performance -- in terms of behaviours that are enabled by ``signs'' in the environment and activated by the decision making process.  In the simulation the agent remembers sequences of tokens representing which behaviours were enabled/activated.  One might want to say that the token ``Wwex'' represents the activation of the go west behaviour given the choice of go west, go east, or eating. The token is however not a \textit{symbolic} representation in that it ``points to'' something internal - there is an explicit causal relationship between the appearance of the token where it appears, and the state of the behaviours it represents.  There is no abstraction, no reference to things in the world, and a statement in terms of such representations cannot be true or false.

In deed in the simulation there are no goals -- no ``image-like representation of a future state of the world''~\cite{tomasello22}.  The roomba remembers that one time it was doing sleeping behaviour and, when it stopped doing that, it went North, then West, then North, East, and then did Eating.  And in classic AI it is easy to see how means-ends reasoning might start with the goal of eating, and plan a sequence of (activations of) behaviours.  Rather than thinking about the state of the agent as hungry and formulating a goal with a plan to achieve mitigate it, the mechanism simply prioritises eating. When the autocomplete mechanism pulls up a memory that led last time to eating, the system favours that path and chooses the next action based on that. Autocomplete over memories is more a means of ``seeing the future'' rather than planning.

So let's call this method of action selection Type 1 thinking, and contrast it with Type 2 thinking in which there are symbolic representations of beliefs about the world and plans are formed to achieve goals.  The claim is that Type 1 thinking is how animals and infants reason about action.  We adults can use Type 2 thinking - which is more powerful of course - and we assume that animals and children think just like us, but poorly.
Object permanence is a classic.  Infants at a certain age will exhibit surprise if a red ball goes in one end of a cardboard tube and a blue one comes out the other end. I also would be surprised if I see an ambulance go past one window, and not past the other.  I explain that expectation with the (scientific) knowledge that I expect the object to remain in existence and have to go somewhere.  Type 1 reasoning however can also explain it. The blue flashing light attracts my attention in the first window and, in the past, that has always been followed by my attention being drawn to a blue flashing light in the other.  I don't need object permanence to detect the correlation, and I don't need the scientific explanation to justify my expectation -- no matter how justified the scientific explanation is.

\section{ Speech and Language } \label{s3}
The above provides a model of ``prelinguistic thought'', what about preintellectual speech?

Reasoning with symbolic representations has a very long tradition that is often seen as ``peaking'' in the west with Witgenstein's Tractatus~\cite{WgsnT}.  Witgenstein is however most famous for moving on and in the Blue and Brown Books~\cite{WgsnBB} he has an example in which a brick layer calls out to his labourer ``Brick!''.  The brick layer, Witgenstein points out, is not saying ``There is a brick'' in the same way as a robber might say to a colleague ``Police!'', or a robot might say to itself ``BRICK\_3''.  The brick layer is \textit{performing} a social act and communicating the need for another brick.  It is a call to action - a performance that is expected to be significant to another human being.  Austin~\cite{austin} formalizes this and demonstrates that the performance does not necessarily involve a reference.  To knight someone or perform a marriage ceremony, nothing physically changes but people's expectations change.  What Vygotsky's group discovered was that people without schooling - (ex) peasants in isolated communities basically - were using language as a means of coordinating action, and might be doing it without the need for referential meaning.  Consider the following vignette: 
\begin{quote2c}
An illiterate peasant from a remote village was shown a hammer, a saw a log, and a hatchet and asked to tell which of the objects could be called by the same word. At first the peasant rejected the very idea of a common name, arguing that you cannot call different objects by the same name. When prompted by the suggestion that ``one fellow said that the hammer, saw and hatchet are alike'', the subject responded that the log also belongs in the same group.  When the experimenter suggested that hammer, saw and hatchet are tools, the subject responded: ``Yes, but even if we have tools, we still need wood - otherwise, we can't build anything''.' [\cite{kozulin90} page 131].
\end{quote2c}
The conclusion was that children, apes, and ``primitive'' societies could get by very well with reasoning in terms of \textit{practices}, and their language was grounded in those. Rather than the world being full of things which could be referred to, the world for a peasant was full of work to be done. The idea that ``intelligence'' is not in heads but rather a cultural artefact is discussed elsewhere~\cite{wallis24} but here the point is that some of these practices involve vocalisation.  In apes and young children these vocalisations ``broadcast'' emotional state without an intention to communicate. Upon hearing that Jim has been scared by a cheetah, a vervet monkey will run to the nearest tree~\cite{vervetCalls} but that does not mean the monkey is working with symbolic representations of cheetahs.  Smoke does not mean fire; smoke is just smoke. Jim's high pitched screech doesn't mean cheetah; it is just a high pitched screech. What Jim is doing is \textit{performing} a speech act but it is only a speech act because others notice it, and figure out that such events correlate with the initiation of a social practice - the practice of running away from cheetahs by climbing trees. 

The proposal is that LLMs produce impressive conversational output with exactly the same mechanism. Although we believe we are being told that glue is good on pizza~\cite{gluePizza}, the LLM is performing the practice of convincing you that glue would be good.  It would of course be anthropomorphising to say the machine wants to convince you more than it wants you to have a good experience with your pizza, but in truth the mechanism just \textit{does} convincing with words.  Making a good pizza is an embodied practice in the world, but the LLM has been trained on the social practice of convincing.  We humans are convinced, and our self deceiving explanation of what is happening is that the machine \textit{understands} what it is saying.  Ask Google in 2026 what a ``CJUL53LOX'' is, and I get the statement ``It is a time-of-flight sensor that can measure distances from 200 to 2000mm..''.  This text is of course a description found on the internet, but the machine recognises this text as the most appropriate for the social practice of convincing.  The descriptions chosen are often extremely appropriate but then human generated descriptions are not objective truth produced in isolation of the practices surrounding them.  Birds of prey on Wikipedia nearly always contain some mention of colouring which is great if you're trying to identify an unknown bird. Descriptions of attack helicopters never mention a colour.

\section{ Prelinguistic thought meets preintellectual language use: the ZPD} \label{s4}

We see a child, an ape, a cat, or a lizard, performing some task and think that it thinks about the task much as we do~\cite{tomasello22}.  Vygotsky and his coworkers found that prelinguistic thought and preintellectual speech gradually merge to become ``scientific''.  The proposal is that the completion mechanism is what lizards, cats, apes and toddlers are using when they do reasoning without language -- without symbols.  Cats, apes, and toddlers also perform preintellectual vocalisations, but these are performative rather than referential.
While we tend to treat the performative side of speech and language as a side issue the proposal is that interaction is the seat of cultural intelligence.  There is nothing natural about scientific thinking - it needs to be taught to successive generations and interaction is the site where this happens.
The argument is that the human \textit{Interaction Engine}~\cite{lev2006,Heesen22} has specific features that enable the transfer of ways of thinking. When an adult and child are in the zone - in the ZPD - the child is trained to think like one of us.

Consider what is needed in this vignette: Amara wakes in the morning and the first word she says is ``duck''. If she is using Type 1 thinking exclusively she is not asserting the existence of ducks.  But if the presence of ducks Amara gets excited, and the last time she was excited by the presence of a duck someone said ``ducks'' then the sound is very likely to be associated with the experience. Further, if Grandad was the one who said ``ducks'' last time, and it is Grandad waking her in the morning, then her auto complete mechanism is likely to call up the sequence (image of) Grandad ... excitedly experiencing duckness. Translated into Type 2 thinking, Amara remembers that being with Grandad is often succeeded by the (very pleasurable) phenomena of being with ducks by the pond. None of Amara's thoughts at this stage need be referential.

Grandad on the other hand is an enculturated adult who does Type 2 thinking. He works hard to \textit{account for}~\cite{Wallis05} his conversational partner's utterances and concludes that Amara wants to go visit the ducks in the pond.  Amara might have said ``uck'', ``dac'', ``yuk'' or any number of things that rhyme with ``duck'' but Grandad, being addressed, is going to interpret his interlocutor's utterance as meaning something.  Amara might have been attending to her full nappy and said ``yuk'' but the noise she made leads to going to visit the ducks in the pond -- which to her is not a bad thing.  If that is not what she wants to do, then Grandad will be interactively corrected, but the lesson has already been learnt -- Amara and Grandad are in the Zone of Proximal Development and Amara learns that the noise she made will \textit{systematically} result in a trip to the ducks. Hopefully her pronunciation of ``yuk'' will at some point correlate with the practice of a nappy change.  The process is systematic because Grandad is always going to interpret her utterances as meaning \textit{something}, but learning that ``duck'' is a reference to the things in the pond will come later.

Developments in the ZPD don't stop there.  When Amara says ``ducks!'' Grandad thinks she \textit{wants} to go see the ducks.  It is possible to say in English that if Amara visited the ducks then she would be very happy -- which is a more accurate way of describing how the mind works by the completion model -- but it is generally thought people use their ``theory of (other) mind'', or ToM, to \textit{account for}~\cite{Wallis22} the odd things their conversational partner say.  Hutto (care of Gallagher~\cite{gallagher2020}) has proposed that, rather than reasoning about other minds, we ``directly perceive'' the intentions of others.  Rather than seeing a woman at a bus stop and reasoning about why she would stand there, what we see is someone waiting for a bus.  Along side ``Theory Theory'', and ``Simulation Theory'' as models of ToM, Pafla adds ``Direct Perception''~\cite{Pafla24}.  I propose here that glorified autocomplete as a form of case-based reasoning as described above is a model of direct perception.

\section{Pre Intellectual Speech (Recognition)} \label{s5}

Word Error Rate has been a very useful tool for developing the tools that we have developed, but there is perhaps an assumption built into the measure that there are word-like thoughts in heads, and there is a mapping between sound and thoughts. The proposed alternative is that thoughts are constructed from sounds through interaction.
The first is consistent with Tomasello's claim that intelligence (as we know it) came first, and social skills were built on that~\cite{tomasello22}. However marmosets are not thought to be the most intelligent of primates but they are highly social and indeed highly vocal~\cite{BurkEtal22}. The callitrichids (marmosets, tamarins ..) have large call repertoires and form call combinations. They engage in turn taking, and adjust call structure to account for environmental and social factors.  There are call 'dialects,' infant vocalisations differ from those of adults, and infants learn \textit{and are trained} to sound like adults in what looks like the callitrichid version of ZPDs. The extensive analysis of callitrichid interaction is of course done by scientists who think ``scientifically'' about things like object permanence and ToM.

But how much of the documented marmoset behaviour can we explain using a completion model of memory based reasoning and statistical correlation?  Marmosets ``show a coupling of the phonatory and articulartory system resulting in a speech-like bi-motor rhythmicity with oscillations that are synchronized and phase-locked at a 3--8Hz theta rhythm'' (Risueno-Segovia and Hage 2020, c.f. Burkart et al~\cite{BurkEtal22}). This looks more like signal processing than word recognition and is perhaps the correct way to describe human-human interaction in the ZPD.

In a similar spirit from an entirely different research tradition, consider my interaction with the slide whistle and bellows in Figure~\ref{whistle}.  As an ethnographic study~\cite{garf67, tHave99} I can report that this device has a ``biometric rhythm'' with which my speech synchronises.  It is, of course, because the device ``runs out of breath'' and so its ``vocalisations'' have a rhythm I detect and synchronize to.  I of course (using scientific thinking) know why it behaves as it does, but that doesn't alter the visceral experience of interacting with something with a rhythm that I notice. I synchronise with it, but what if I didn't? What would that \textit{mean?}  Should the whistle syncronise with me? And what would that mean?

\section{Conclusion}

LLMs don't think like us, and neither do marmosets or toddlers.  Vygotsky's claim was that Type 2 thinking is a cultural product and learnt by the infant in a ZPD.  For ZPDs to work requires special features from the human interaction engine, but studies of interaction in that space largely assume type 2 thinking.
So how would we study non symbolic communication?  How would we study the ``speech'' of marmosets and human interaction with artifacts like the whistle? Alarms and ring tones work, but can we do more with their (non referential) semantics?   For vocal interaction the \textit{how} looks to be primarily signal processing.  The \textit{what} is communicated is perhaps anthropology, conversation analysis, and ethnomethodological studies.

\bibliographystyle{IEEEtran}
\bibliography{~/mybib.bib}

\end{document}